%% file: main.tex
\documentclass[journal]{IEEEtran}
\usepackage{amsmath,amsfonts}
\usepackage{algorithmic}
\usepackage{algorithm}
\usepackage{array}
\usepackage[caption=false,font=normalsize,labelfont=sf,textfont=sf]{subfig}
\usepackage{textcomp}
\usepackage{stfloats}
\usepackage{url}
\usepackage{verbatim}
\usepackage{gensymb}
\usepackage{graphicx}
\usepackage{cite}
\usepackage{multirow}
\usepackage{cleveref}
\usepackage{amsmath}
\usepackage{mathtools}
\usepackage{upgreek}
\usepackage{balance}
\usepackage{xcolor}

\crefformat{section}{\S#2#1#3} 
\crefformat{subsection}{\S#2#1#3}
\crefformat{subsubsection}{\S#2#1#3}

\begin{document}
\title{Transforming Future Data Center Operations and Management via Physical AI}

\author{
\IEEEauthorblockN{
    Zhiwei Cao\IEEEauthorrefmark{1},
    Minghao Li\IEEEauthorrefmark{1},
    Feng Lin\IEEEauthorrefmark{1},
    Jimin Jia\IEEEauthorrefmark{1},
    Yonggang Wen\IEEEauthorrefmark{1},
    Jianxiong Yin\IEEEauthorrefmark{2},
    Simon See\IEEEauthorrefmark{2}
}\\
\IEEEauthorblockA{
    \IEEEauthorrefmark{1}
    Nanyang Technological University, Singapore
}
\IEEEauthorblockA{
    \IEEEauthorrefmark{2}
    NVIDIA AI Technology Center
}
}
\markboth{Journal of \LaTeX\ Class Files,~Vol.~14, No.~8, August~2015}%
{Shell \MakeLowercase{\textit{et al.}}: Bare Demo of IEEEtran.cls for IEEE Journals}

\maketitle

\input{abstract}
\IEEEpeerreviewmaketitle

\input{body/sec1-intro}
\input{body/sec2-related-work}
\input{body/sec3-overview}
\input{body/sec4-case-study}
\input{body/sec5-future-work}
\input{body/sec6-conclusion}

\balance
\bibliographystyle{IEEEtran}
\bibliography{main}

\end{document}

%% file: abstract.tex
\begin{abstract}
Data centers (DCs) as mission-critical infrastructures are pivotal in powering the growth of artificial intelligence (AI) and the digital economy.  The evolution from Internet DC to AI DC has introduced new challenges in operating and managing data centers for improved business resilience and reduced total cost of ownership. As a result, new paradigms, beyond the traditional approaches based on best practices, must be in order for future data centers. In this research, we propose and develop a novel Physical AI (PhyAI) framework for advancing DC operations and management.  Our system leverages the emerging capabilities of state-of-the-art industrial products and our in-house research and development. Specifically, it presents three core modules, namely: 1) an industry-grade in-house simulation engine to simulate DC operations in a highly accurate manner, 2) an AI engine built upon NVIDIA PhysicsNemo for the training and evaluation of physics-informed machine learning (PIML) models, and 3) a digital twin platform built upon NVIDIA Omniverse for our proposed 5-tier digital twin framework. This system presents a scalable and adaptable solution to digitalize, optimize, and automate future data center operations and management, by enabling real-time digital twins for future data centers. To illustrate its effectiveness, we present a compelling case study on building a surrogate model for predicting the thermal and airflow profiles of a large-scale DC in a real-time manner. Our results demonstrate its superior performance over traditional time-consuming Computational Fluid Dynamics/Heat Transfer (CFD/HT) simulation, with a median absolute temperature prediction error of 0.18 °C. This emerging approach would open doors to several potential research directions for advancing Physical AI in future DC operations.
\end{abstract}

%% file: body/sec1-intro.tex
\section{Introduction}
The data center (DC) industry is rapidly growing to meet the increasing demands for cloud computing and storage services. This growth trend is evident in the DC's rising scale and power consumption. The DC scale can be measured by its size and the equipment quantity. For example, a hyperscale DC typically spans thousands of square feet and hosts thousands of information technology (IT) devices \cite{hyperscale}. Such growing scale elevates the power required to operate these IT devices and their associated cooling systems. According to the latest report from the International Energy Agency (IEA), the DC industry is projected to consume up to 1050 TWh of energy by 2026, compared to 460 TWh in 2022 \cite{iea2024}. Another emerging trend involves the development of graphics processing unit (GPU) DCs tailored to artificial intelligence (AI) training and inference purposes. For GPU DCs, the per-rack power density can reach up to 50kW or 100kW \cite{gpu-dc}.

The growth of DCs' scale and power usage brings management challenges in reducing costs while ensuring system reliability. On the one hand, the cost mainly comes from capital (CapEx) and operational (OpEx) expenditures. The CapEx of building a DC with 100k B200 GPU ranges from 4.5 to 5 billion US dollars and the annual electricity OpEx can reach 400 million \cite{price-B200}. Given the high cost of developing and operating a DC, efficient and economical solutions are highly desirable. On the other hand, the reliability of a DC is often measured by its uptime, which determines the business continuity. The reliability is related to multiple cyber and physical factors. In particular, the physical factors, like unplanned power outages, matter greatly in ensuring a DC's reliability \cite{reliability}.

To address the challenges, the industry and academia are jointly leading the way to advance DC operations. The DC operations maturity model indicates that most industrial DCs function at a reactive level \cite{maturity}. This level adopts the data center infrastructure management (DCIM) system to monitor the DC states and employs feedback control. For instance, the proportional-integral-derivative (PID) controller is adopted by the air conditioning units to keep the data hall temperature around a predetermined setpoint. However, these controllers solely focus on maintaining temperature without incorporating other objectives, like power consumption. In addition, most current DCIM systems lack precise prediction capabilities, which are needed for conducting a variety of what-if analyses, such as determining if increasing certain temperature setpoints could decrease power usage without leading to server overheating. 

To evolve DC operations to a proactive and adaptive level, researchers have explored artificial intelligence (AI)-powered solutions. To date, several AI-powered optimizations have been applied in data center management, including IT resource optimization, cooling control optimization, battery predictive maintenance, etc. For example, AI can analyze massive historical data regarding the hardware operating conditions and data center environment to predict potential hardware failures before they occur ~\cite{mohapatra2023large}. It minimizes the system-critical risks and prevents unexpected downtime to secure stable data center operation. In computing system management, several AI-based resource management and workload orchestration systems have been proposed recently ~\cite{mao2019learning}. AI learns from historical data and dynamically adjusts the resource allocation, scheduling, and prioritization of critical workloads and services hosted in data centers to improve operational efficiency, resource utilization, and sustainability. In the physical facility management system, AI optimizes data center energy consumption by analyzing data on power usage, temperature, and workload characteristics. It can identify opportunities for energy savings, such as adjusting cooling system setpoints ~\cite{LyaSafe2023, wang2022toward}, to significantly reduce operation costs and boost overall sustainability. 

Although AI-empowered solutions have shown promising performance in various aspects of DC operation, two significant challenges hinder their practical adoption: 1) data scarcity and 2) strict operational regulation compliance guarantee. Data centers are usually operated according to industrial best practices, indicating that dynamic online data is generally unavailable. Without data covering sufficient operation conditions, AI-based solutions fail to learn complex system dynamics, and their generalization cannot be guaranteed. As DCs are risk-sensitive entities with strict reliability requirements, the solutions without regulation compliance guarantee face significant challenges in real-world deployment. As evidence, it is reported by Uptime Institute that only 19\% DC operators believe that AI-based solutions will replace human staff in DC operations shortly \cite{uptime-2022}. Thus, fully autonomous DC operation remains challenging with the existing paradigm.

\begin{figure*}[t]
    \centering
    \includegraphics[width=.85\textwidth]{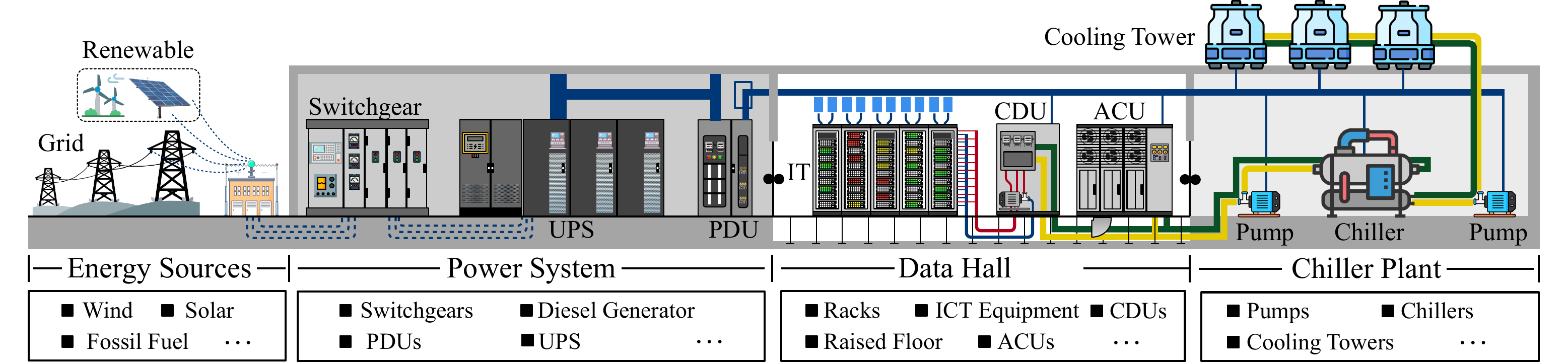}
    \caption{Illustration of the typical DC physical infrastructure, which consists of the IT system, the cooling system, and the power supply system. The IT system hosts various workloads, ranging from traditional cloud services, networking, and storage workloads to emerging AI workloads. To support the high-density AI workloads, hybrid cooling systems (liquid cooling + air cooling) will be widely adopted in the future. To mitigate the carbon and energy footprint, mixed energy systems that integrate multiple energy sources including traditional grid electricity and green energy are prevailing.}
    \label{DC Overview}
\end{figure*}

In this article, we envision a novel physical AI (PhyAI) solution to fully unleash AI's potential in DC operations. Physical AI refers to integrating AI with physical systems and processes, making AI capable of \textit{perceiving}, \textit{understanding}, and \textit{actuate} in the real (physical) world \cite{PhysicalAI}. It extends the current AI by introducing \textit{physically-based} data in the training process to let AI learn the underlying physical rules of the world. The data is generated via high-fidelity computer simulation, serving as the source and foundation for PhyAI training. In particular, we first develop the virtual 3D environment with \textbf{NVIDIA Omniverse} \cite{Omniverse}, a platform integrating cutting-edge rendering technologies and generative physical AI models. A scene generator is developed to generate various data center physical layouts and physical system parameters (e.g., the cooling system setpoints). Our in-house physically-based simulator generates high-fidelity simulation data. The AI engine then leverages the data to train the AI model that simulates real-world DC scenarios rapidly under varying parameters with \textbf{NVIDIA PhysicsNeMo} \cite{PhysicsNeMo}. To model the complex DC environment, we adopt advanced physics-informed machine learning (PIML) techniques for data-efficient and interpretable learning. Specifically, PIML injects physical knowledge into ML models by adding penalty terms in the learning objective or curating physics-inspired model architecture to improve its generalization capability, especially in small-data regimes ~\cite{karniadakis2021physics}. With this simulation capability, we adopt the advanced control optimization framework based on Deep Reinforcement Learning (DRL) \cite{franccois2018introduction} and learning-based MPC \cite{hewing2020learning} to obtain performative agents to operate a physical DC. The intelligent agent first explores the virtual environment to gather extensive synthetic data. Subsequently, a hybrid dataset that contains both online and synthetic data is curated. Lastly, it learns to efficiently and reliably operate a large-scale physical DC from these data. We further conduct a case study on the application of PIML to achieve \textit{real-time} thermodynamics simulation of a large-scale production DC to validate the proposed framework. The contributions of this article are summarized as follows.
\begin{itemize}
    \item To the best of our knowledge, it is the \textit{first} article that systematically introduces PhyAI and its application on DCs. To begin with, we first review the DC physical infrastructure. Then, we discuss the basics and building blocks of a PhyAI system. 
    \item We present a principled framework for intelligent DC operations with Physical AI. It leverages the physically-based high-fidelity simulators to help AI agents understand complicated DC physical infrastructure and gradually learn to operate DCs autonomously. Potential applications are also extensively discussed thereafter. 
    \item We take the rapid thermodynamics simulation of a large-scale production DC as the case study, in which we develop a PIML model to achieve \textit{real-time} inference of the thermal profiles of the DC without significantly comprising the prediction accuracy. To the best of our knowledge, it is the \textit{first} case study that demonstrates the effectiveness of NVIDIA PhysicsNemo in modeling real-world physical DCs.
\end{itemize}


%% file: body/sec2-related-work.tex
\section{Preliminaries}\label{Preliminaries}
In this section, we first provide an overview of the physical infrastructure of a data center. Subsequently, we introduce the basics of a physical AI system and its building blocks. 

\subsection{Overview of Data Center Physical Infrastructure}
Data centers (DCs) comprise three fundamental interconnected systems: the information technology (IT) system, the cooling system, and the electrical power supply system, as illustrated in \cref{DC Overview}.  The IT system, responsible for executing computational workloads, is the primary energy demand source.  These workloads can be broadly classified into real-time services and batch workloads.  Critically, batch workloads, such as machine learning model training and big data analytics, exhibit a degree of delay tolerance, enabling scheduling optimization within defined deadlines. This characteristic presents an opportunity for energy efficiency gains.

The cooling system is essential for dissipating the thermal energy generated by the IT infrastructure. In a conventional air-cooled DC, Air Handling Units (ACUs) facilitate airflow, supplying cool air to IT equipment inlets and extracting heated air from outlets. Recently, with the increasing rack power density driven by AI workloads, direct-to-chip (D2C) liquid cooling has emerged as a promising candidate as it provides significantly more cooling capacity compared with traditional air-based cooling. The coolant distribution unit (CDU) is installed to distribute the coolant to IT servers via a pipe network. The coolant will be pumped inside IT servers and cool down electronic devices via heat transfer. A chiller plant supplies cold water to ACUs and CDUs to remove the extracted heat and maintain the desirable air and coolant temperature, ensuring the reliable operation of the IT system.

The electrical power supply system provides power to both the IT and cooling systems. A microgrid architecture integrates diverse power sources, including grid power and on-site renewable energy generation.  The Uninterruptible Power Supply (UPS) provides a critical buffer against power interruptions, ensuring operational continuity.  Power is distributed from the UPS through Power Distribution Units (PDUs) to individual server racks and cooling devices. This integrated power system is designed for reliability and scalability to meet the demands of high-performance computing.

\subsection{Physical AI: Basics and Building Blocks}
In this section, we present the basics of Physical AI and its three building blocks, i.e., the virtual 3D environment, the high-fidelity physically-based simulators, and the engine to train predictive and prescriptive AI models.
\subsubsection{Virtual 3D Environment}
Physical AI starts with constructing a high-fidelity, physically-based environment to generate visually realistic and physically plausible synthetic data to train the intelligent agent that can understand the underlying physical rules.
Current approaches leverage 3D digital twin platforms to provide the foundation for constructing and managing complex virtual environments that accurately represent real-world scenarios.
It supports integrating data from various sources, including IoT sensors, to create dynamic and interactive virtual 3D scenes.
Accurate descriptions of objects and their specifications, combined with high-quality graphics rendering, are essential for reflecting real-world physics.
In this regard, NVIDIA Omniverse enables developers to easily leverage Universal Scene Description (USD) and RTX rendering technologies to build and visualize these 3D environments.
USD unifies the representation of 3D objects with their inherent physical properties, enabling a more comprehensive and realistic simulation of the real world.
It empowers developers to create more accurate and interactive digital scenes, where objects not only exist visually but also behave according to the laws of physics.
To simulate underlying physics, third-party simulators are widely integrated as modular software components, enabling users to customize the platform to their specific needs.

\subsubsection{Physical Simulators}
Physical simulators are essential tools in developing and deploying PhyAI systems. These simulators create virtual environments where AI agents can learn and interact with the physical world, offering a safe, cost-effective, and efficient alternative to real-world experimentation.  Agents can learn complex tasks, such as robot manipulation, locomotion, and navigation, without the risks and costs associated with real-world training. For example, Genesis allows for training robotic locomotion policies in just 26 seconds, significantly accelerating the development process \cite{Genesis}. In addition, simulators can generate large datasets of synthetic data for training AI models. This data can supplement real-world data, improving the robustness and generalization of intelligent agents. Before deploying PhyAI systems in the real world, simulators allow for rigorous testing and evaluation. Different scenarios and environmental conditions can be simulated to evaluate the performance and safety of intelligent agents. Autonomous driving simulators, for instance, enable testing of the systems in various traffic situations and road conditions. Hence, physical simulators offer a safe, cost-effective, and efficient way to train and evaluate AI agents in virtual environments. 

In the context of DC operations, there are several available physical simulators to simulate individual DC sub-systems. For the simulation of IT systems, CloudSim provides a generalized and extensible simulation framework that enables modeling and simulation of the IT infrastructures and application services hosted in a physical DC \cite{andreoli2025cloudsim}. Given a set of workloads, it can simulate the consequence of different resource management and job scheduling policies, providing detailed information on the status of each task as well as the server-level performance metrics such as utilization, power consumption, etc. To simulate the detailed thermal and airflow profiles within a data hall, commercial software, e.g., Cadence Design Systems \cite{CadenceDesignCenterDCDigitalTwin}, is a potential solution as it provides comprehensive modeling capability of the IT servers and cooling equipment. Open source counterparts, e.g., OpenFOAM \cite{jasak2007openfoam}, are available to serve the same purpose as well. For the energy consumption and electrical delivery system, EnergyPlus \cite{crawley2001energyplus}, an open-source simulator, is a powerful tool as it provides comprehensive modeling capability of the HVAC system as well as the electrical delivery system. It can simulate the HVAC system performance under different operating conditions, as well as the net electricity consumption considering the integrated renewable energy generators and the energy storage devices. Thus, it provides detailed energy-related information, which serves as the metrics for cost optimization purposes. 

\subsubsection{AI Engine}
Leveraging the high-fidelity synthetic data from the physical simulators, the AI engine delivers the predictive models and prescriptive agents. Firstly, predictive models are trained to conduct rapid high-fidelity simulations of the physical world. Traditional physics-based simulators solve complex physical problems over discrete meshes, resulting in large-scale nonlinear systems to be solved. Solving such systems consumes a considerable amount of computing resources and the simulation latency is unaffordable in real-time decision-making and intelligent agent training. In this regard, PIML (PIML) approaches are promising as they inject physics, as expressed by governing equations, boundary conditions, and physical simulation data to build high-fidelity surrogate models with extraordinary generalization capability. As a representative, NVIDIA PhysicsNemo provides a useful abstract for complex physical problems and a scalable training pipeline. Besides, prescriptive models that generate executable actions are trained with the synthetic data as well. As the synthetic data contains rich unseen scenarios in the online data, the intelligent agent learns skills safely and quickly through thousands or even millions of interactions with the virtual environment, significantly alleviating the risk-averse mindset of DC operators.  

%% file: body/sec3-overview.tex
\section{Physical AI for Autonomous DC Operations: Framework and Applications}\label{Framework and Applications}
\begin{figure*}[t]
    \centering
    \includegraphics[width=0.85\textwidth]{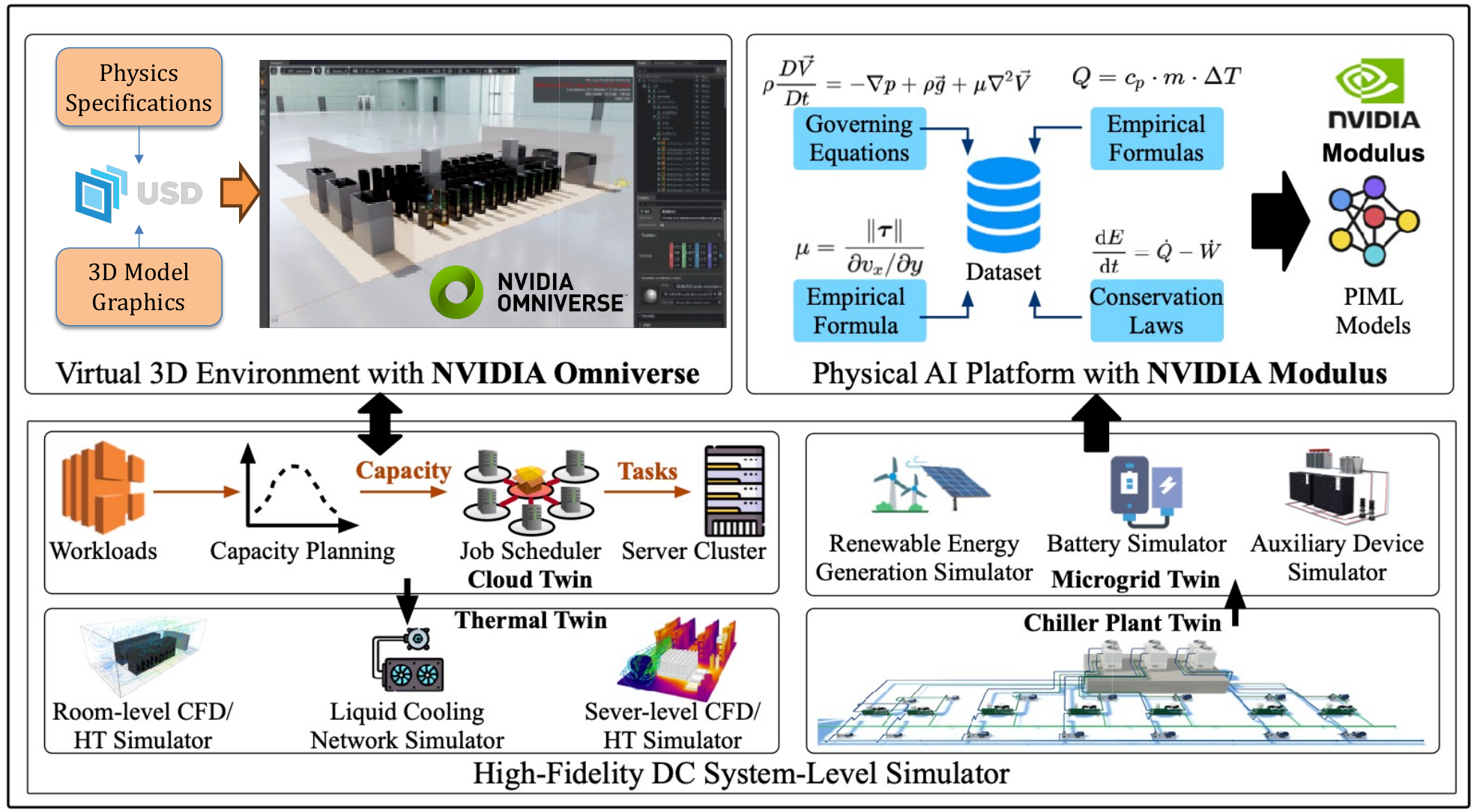}
    \caption{Illustration of the PhyAI-driven system for autonomous DC operations. We build the 3D representation of a physical DC with NVIDIA Omniverse, a powerful platform for constructing and rendering 3D environments with the RTX rending technology. We also enable real-time sensory data visualization and analysis on Omniverse. Built on Omniverse, we develop a bidirectional converter to seamlessly transfer data between our in-house physics-based simulators for system-level DC simulation. The synthetic dataset is fed to the PhyAI engine built on NVIDIA PhysicsNemo, a modern framework for large-scale physics-informed AI model training and inference. The trained PhyAI model is deployed in physical DCs for predictive and prescriptive analysis.}
    \label{PhyAI DC System Overview}
\end{figure*}
In this section, we present the PhyAI framework for autonomous DC operations. Specifically, we first introduce the overall framework design. Then, we articulate several potential applications of PhyAI in DC's \textit{predictive} and \textit{prescriptive} analysis. 

\subsection{Framework Design}
We hereafter introduce the unified framework for designing, training, validating, and deploying PhyAI models, which provide robust and performative AI models to advance DC operations. As shown in \cref{PhyAI DC System Overview}, our framework follows the general PhyAI system design. 

We adopt \textbf{NVIDIA Omniverse} as the digital twin platform to construct a mimic 3D virtual environment. The digital twin encompasses a 5-tier architecture: i) geometric, ii) descriptive, iii) predictive, iv) prescriptive, and v) automation, where each tier builds upon the previous one, progressing from geometry modeling to predictive simulation, then to intelligent decision-making and automated execution. Our virtual environment is implemented as an Omniverse extension. Our extension provides high-resolution 3D scene rendering and simulation visualization for physical DCs. We develop a sim-ready assets library encompassing a comprehensive set of 3D and physics models, including multi-level cooling facilities and heterogeneous IT equipment (ITE) in DC. Instead of using structured make-up language (e.g. JSON, XML) and 3D geometric files (e.g. obj, stl), we adopt universal scene description (USD) as the \textit{unified} file format of our DC digital twin description to store both geometric information and physics specifications for the descriptive and predictive digital twin. USD is an open-source, extensible framework for describing, composing, and simulating 3D scenes. To integrate physics attributes into USD, our extension offers a bidirectional converter that allows to import of any DC scene in the JSON format, editing the DC scene and physical characteristics with Omniverse GUIs, and saving the DC USD scene back into JSON format. Besides, our extension also provides real-time connections to the IoT system and our PhyAI engine. For DC operations, our extension receives streaming and periodic data from on-premise edge servers in physical DCs and forwards them into the PhyAI engine for simulation and optimization. The computing results are then fed into the visualization module built upon the Paraview Python SDK. It eventually converts the numerical results into USD files for Omniverse rendering. We design an interactive 3D interface featuring floating panels, which allows quick what-if analysis by modifying the boundary conditions for any facility and ITE. This interface further supports the validation of control algorithms to achieve the prescriptive digital twin for operation optimization. Ultimately, our digital twin platform enables automation through edge computing frameworks (e.g., AWS IoT Greengrass, NVIDIA Jetson) for direct policy deployment, real-time actuation, and control of IoT devices across facilities.

On top of the virtual environment, we develop a comprehensive physical simulation engine as the synthetic data generator. The simulation engine performs the system-level simulation of a physical DC, providing fine-grained information about the DC under different operating conditions. The entire simulation workflow starts with the IT system. Upon receiving the workload information, the IT simulator conducts resource allocation and job scheduling to determine which server to place the workload on. The power consumption model inside the IT simulator calculates the detailed server-level power consumption accordingly. The server-level power consumption information combined with the data hall cooling system setpoints will be the input to the data hall thermodynamics simulator to compute \textit{high-fidelity} thermal and airflow profile. The thermodynamics simulator is built upon OpenFOAM. We further develop adaptive meshing and GPU-based acceleration capability to speed up the physical simulation. The simulated thermal profile and the IT system power consumption are then fed into the energy simulators to compute the energy consumption of the physical facility of the DC. We built the energy simulator with EnergyPlus, which provides the HVAC and electrical delivery systems' modeling capability. In the electrical delivery system, we incorporate renewable energy generators (e.g., photovoltaic panels and wind turbines) and battery devices to consider energy consumption with different renewable energy integration levels and battery operation policies. We follow a \textit{modularized} principle to design the simulation engine, indicating that each simulator can work individually, offering great flexibility for customization. 

With the synthetic data generated by the simulation engine, we develop an AI engine based on \textbf{NVIDIA PhysicsNeMo}, providing a scalable pipeline for physical AI model training, validating, and deploying. The AI engine delivers high-quality surrogate predictive models and decision-making agents. Large-scale physical simulations typically consume a huge amount of computing resources and they take a long time. To speed up the decision-making agent training process, we leverage the state-of-the-art PIML techniques to train fast and robust PIML-based surrogate models that can be simulated in \textit{real time}. Using the accelerated simulation capability, we further train decision-making agents that learn to optimize certain performance metrics. With the fast and accurate surrogate predictive models, the decision-making agents are trained with a learning-based framework, in particular, the DRL and learning-based MPC. Intelligent agents explore the virtual DC environment and receive feedback from the physics-informed surrogate models in real-time. Via extensive trial-and-error, autonomous agents accumulate massive amounts of experience data, and then gradually learn the underlying physical rules. Eventually, they can adapt to new situations and unforeseen challenges appropriately, preparing them to operate in real DCs.

Once AI models are fully trained in the virtual environment, they are deployed to a physical DC to perform complex tasks like computing resource management, cooling system setpoint adjustment, and energy storage scheduling. The real physical DC generates the corresponding response, which will be collected and visualized via the Omniverse platform. The online sensory data will be stored in the database, and then be used to calibrate the AI models and augment the synthetic dataset. 

\begin{figure*}[t]
    \centering
    \includegraphics[width=0.85\textwidth]{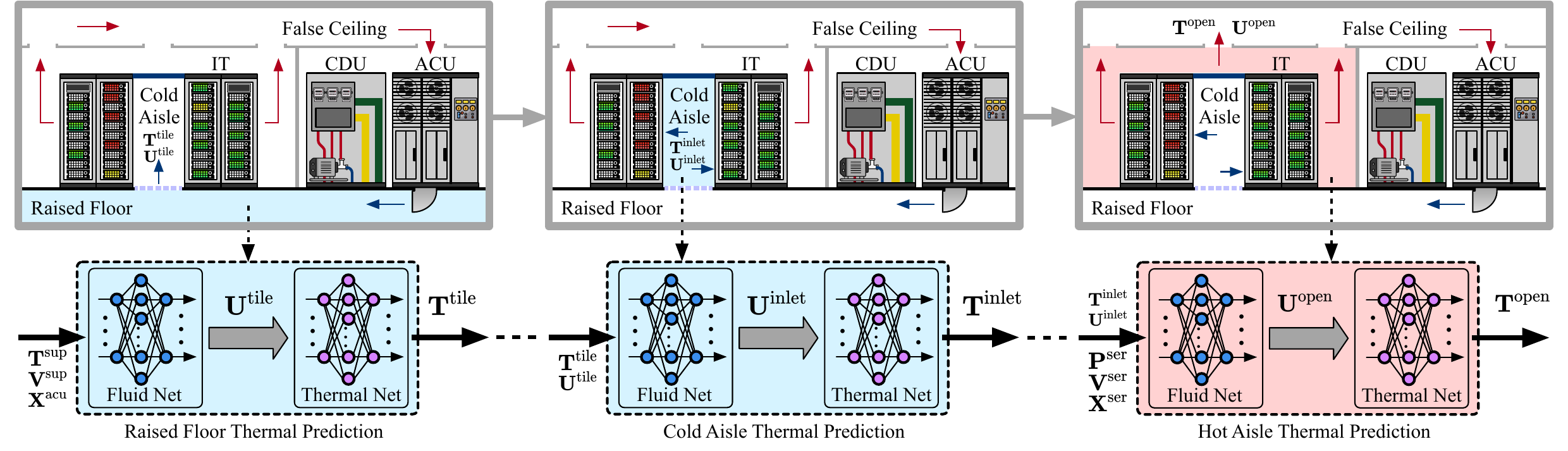}
    \caption{Illustration of the proposed physics-informed model architecture. We conduct \textit{domain decomposition} and split the computation domain into three parts, i.e., the raised floor area, the cold aisle, and the hot aisle as the flow and thermal profiles in the three areas are significantly distinct. In addition, \textit{separate modeling }of the thermal and airflow is adopted as the fluid field and thermal field are loosely coupled in the context of the room-level cooling process. We first simulate the flow field and then use the simulated flow field to infer the thermal field according to the governing equations. To ease the training, we first train the fluid network and then freeze it to train the thermal network.}
    \label{Surrogate Model Architecuture}
\end{figure*}
\subsection{Potential Applications}
In this section, we introduce the applications of PhyAI in DC \textit{predictive} and \textit{prescriptive} analysis. 

\subsubsection{Physical AI for Predictive Analysis}
High-fidelity physical simulators are powerful tools to perform predictive analysis for a physical DC, but they are likely to be computationally expensive, requiring powerful hardware and efficient algorithms. A considerable amount of research has proposed leveraging deep learning techniques to approximate the solution from the high-fidelity simulator in a data-driven manner. Although these approaches significantly accelerate high-fidelity simulation, the black-box nature makes them less interpretable and might cause severe extrapolation errors when evaluated on unseen scenarios. 

PIML techniques are emerging as a strong candidate for future predictive analysis of physical DCs. On top of traditional data-driven AI, PIML integrates physical (domain) knowledge into ML models, yielding more accurate and physically plausible results. In the context of DCs, PIML can be applied to establish the dynamical system of the physical facility. For instance, a PIML-based surrogate model can infer the high dimensional temperature and airflow profiles in real-time, substantially accelerating the computational-intensive CFD/HT simulation \cite{cao2022reducio}. Besides, the surrogate model can assimilate the online sensory data and then provide the calibrated CFD/HT simulator parameters \cite{wang2020kalibre}. For the energy system, the PIML-based approach has shown superior accuracy over traditional DL models in identifying the complicated dynamics of the HVAC system of a physical DC, even in the small data regime \cite{Cao2023Adaptive}. One potential next step is extending the PIML paradigm to other devices such as batteries, thermal storage tanks, renewable energy generators, etc. With these models, a comprehensive PIML-based predictive engine is constructed, capable of conducting system-level real-time and accurate predictive analysis for a physical DC.

\subsubsection{Physical AI for Prescriptive Analysis}
With the predictive analysis capability, DC operators can design intelligent agents to conduct prescriptive analysis to advance operation efficiency. In this regard, two algorithmic frameworks are considered, i.e., DRL and learning-based MPC. RL rewards an intelligent to completing certain tasks. It typically involves training a policy that maps observable states to executable actions by trial and error in the environment. The aforementioned predictive engine can serve as a virtual environment where the agent can safely explore and learn at a low cost. The major drawback of DRL is its limitation in handling complex constraints. Learning-based MPC, on the other hand, provides a natural way to handle the operating constraints by formulating them into a mathematical program. Its key is to learn an accurate and easy-to-evaluate system dynamical model, where PIML is applicable. One limitation of MPC is that it requires solving complicated optimization online, which incurs considerable latency. 

For DC applications, these two frameworks are widely adopted and have shown promising results in improving DC operation efficiency. The DRL-based approach has been adopted in the resource provisioning and scheduling of large-scale computing clusters, outperforming traditional heuristic-based approaches significantly \cite{LyaSafe2023}. Besides computing cluster management, DRL approaches can be applied to managing the physical facility of a DC. For instance, recent research has shown that a DRL-based agent can dynamically adjust the setpoints of the HVAC system to improve its energy efficiency. The feasibility of joint optimization of IT and facility is also explored and further energy savings are demonstrated \cite{Zhou2023JointIT}. For the MPC-based approach, the recent study shows that an MPC-based capacity provisioning scheme can reduce the carbon footprint of a large-scale DC by 27\% while complying with the system operating constraints \cite{Cao2023Adaptive}. Another study leverages online data to identify the dynamics of the cooling system and designs an MPC-based agent to dynamically adjust the setpoints of the HVAC system, showing at most 40\% cooling energy savings \cite{lazic2018data}. In summary, with the aid of the high-fidelity physical simulator, both learning-based prescriptive analysis frameworks are applicable for optimizing different systems of a physical DC as well as conducting holistic optimization by considering the complicated trade-off between different systems.

%% file: body/sec4-case-study.tex
\section{Cast Study: Rapid CFD/HT Simulation with PIML for Proactive DC Thermal Management}\label{Case Study}
To validate the effectiveness of the proposed framework, we present a case study on PIML-based proactive thermal and airflow management for a large-scale production DC. 

\subsection{Motivation}
Modern DCs are typically equipped with Data Center Infrastructure Management (DCIM) systems, which facilitate the monitoring of system states and support operators in identifying potential risks such as unplanned server shutdowns and localized thermal hot spots. However, as the scale and complexity of DCs continue to grow, such reactive monitoring approaches become increasingly inadequate for anticipating failures. Consequently, there is a critical need to augment DCIM systems with predictive models that are both accurate and timely.

Computational Fluid Dynamics and Heat Transfer (CFD/HT) simulation is a widely adopted technique for predictive thermal analysis in physical DCs. By solving the Navier–Stokes equations in conjunction with the energy balance equation, CFD/HT models can generate fine-grained airflow and temperature distributions. When properly calibrated, these models can achieve temperature prediction errors of less than 1°C \cite{wang2020kalibre}. Despite their accuracy, the high computational overhead associated with CFD/HT simulations poses a significant limitation for real-time or near-real-time thermal forecasting. For instance, when applied to hyper-scale DCs with fine mesh resolution, one simulation costs hours even days to complete. As a result, CFD/HT simulations are primarily utilized during the prototyping and design phases of DC development.

\begin{figure}[t]
    \centering
    \includegraphics[width=0.46\textwidth]{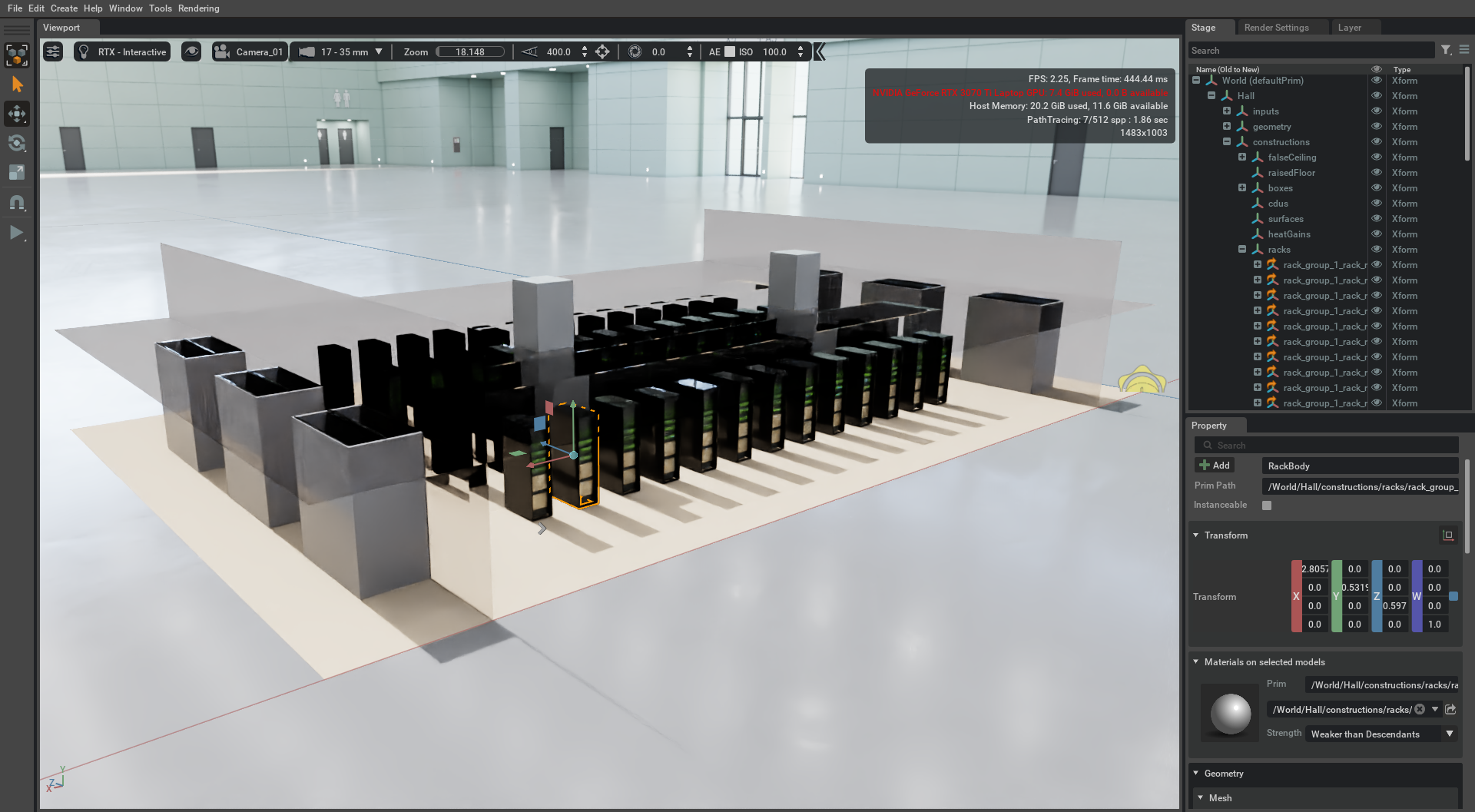}
    \caption{Illustration of the layout of the considered DC with NVIDIA Omniverse. The data hall contains 4 rows of racks, with a total rack number equaling 60. 317 servers are installed in the racks. 6 ACU units are installed to provide cold air. Hot aisle containment is equipped to improve cooling efficiency.}
    \label{DC Layout}
\end{figure}
\subsection{Problem Settings}
In this case study, we explore a PIML-based approach for accelerating DC CFD/HT simulation at scale, shedding new light on future proactive DC thermal management. Specifically, we consider a production DC in Singapore with the data hall area around 600 $\text{m}^2$, as shown in \cref{DC Layout}. The data hall contains 4 rows of racks, with a total rack number equaling 60. 340 servers are installed in the racks, each containing 3-8 servers. The DC adopts a traditional air-based cooling system, and 6 ACUs are installed inside the data hall. In addition, hot-aisle containment is also adopted to isolate cold and hot air and improve air delivery efficiency. Each ACU is specified by its air supply temperature and supply air volumetric flow rate ($\text{m}^3$/s). The configuration of a server includes its power consumption ($\text{W}$) and inlet air volumetric flow rate ($\text{m}^3$/s). Given these configurations, the PhyAI model provides the fine-grained velocity and temperature profile of the data hall in \textit{real-time}. 

To train the surrogate model, we first leverage our high-fidelity room-level CFD/HT simulator to generate a high-quality synthetic dataset covering various boundary condition settings. Specifically, we sample the supply temperature using the Latin Hypercube approach with the range of 18\degree C to 24\degree C.  We adopt the Latin Hypercube approach with the range of 1000 to 2000 $\text{W}$ to sample the server boundary conditions. In total, 500 different scenarios are sampled and simulated with our physical simulation engine. The geometry and mesh of each case is the same, and the number of mesh grids is around 340k. With our optimized GPU-based numerical solver, each case can be finished in around 5 minutes with 1 NVIDIA RTX 4090 GPU. We randomly select 400 cases for training, 50 cases for validation, and 50 cases for evaluation. The model is trained with 4 NVIDIA V100 GPUs with 32 GB memory. 

\subsection{Model Architecture}
We build the physics-informed neural surrogate model upon the state-of-the-art General Neural Operator Transformer (GNOT) since it is scalable and applicable to problems with multiple inputs and arbitrary geometry \cite{hao2023gnot}. Specifically, the model accepts three inputs, i.e., the server input function, the ACU input function, and the query points to be evaluated. The server input contains the inlet and outlet face coordinates, the power consumption, and the inlet volumetric flow rate of each server. The ACU inputs consist of the supply face coordinates, the supply temperature, and the supply volumetric flow rate of each ACU. Concerning the query points, we extract the cell centers from the OpenFOAM simulation results. Three different encoders are adopted for different kinds of input features. The multi-layer perceptron (MLP) is used as the encoder for the server and ACU inputs, which projects the original input function space to the latent space. For the query points, Fourier features \cite{tancik2020fourier} are employed in the encoder to facilitate the model capturing high-frequency components in the solution functions. Once the three inputs are encoded properly, they are processed with the cascaded Transformer blocks with the cross-attention and the self-attention module. Both attention modules are Galerkin-type with \textit{linear} complexity w.r.t. query points to scale the model into the large-scale problem.

We also incorporate several DC-specific designs into the model to improve its performance and ease the training process. Firstly, we conduct \textit{domain decomposition} when training the model as shown in \cref{Surrogate Model Architecuture}. Specifically, the entire data hall is decomposed into three domains, i.e., the cold aisle and the two hot aisles. We first train the model to predict the thermal and velocity profiles of the cold aisle using only the ACU inputs and the query points within the cold aisle. Then, we train the model for each hot aisle separately to predict the thermal and velocity profiles of each hot aisle. With domain decomposition, we reduce the memory and computing overhead needed to train the surrogate model. Meanwhile, the scalability of the modeling approach is significantly enhanced because the parameter space for each model is significantly reduced, alleviating the difficulty of sampling boundary conditions in high dimensional space. Furthermore, we also consider server-level mass and energy conservation to be physical constraints and integrate them into objective functions as soft penalties. It makes the model aware of the underlying physical rules, improving its convergence and generalization. 

\subsection{Evaluation Results}
\begin{figure*}
    \centering
    \includegraphics[width=1\textwidth]{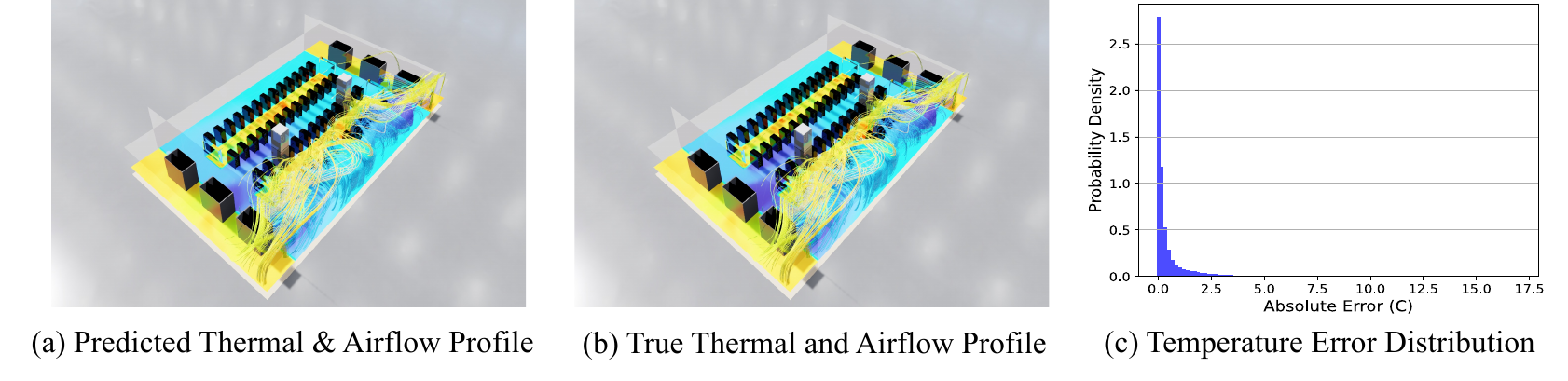}
    \caption{Illustration of the predicted and the true temperature and airflow profile, as well as the absolute temperature prediction error distribution. The airflow profile is visualized with streamlines in Omniverse. It can be seen that the surrogate model produces visually similar thermal and airflow profiles compared with the ground truth. The quantitative result shows that for most areas within the data hall, the absolute temperature prediction error is within 2.5 \textdegree C with a median absolute error of 0.18 \textdegree C.}
    \label{Result}
\end{figure*}

The evaluation result is shown in \cref{Result}, which illustrates the predicted thermal map and airflow streamlines, and the error distribution of the temperature and airflow profile. From the visualization, it can be seen that the surrogate model fields accurate prediction of both temperature and airflow profiles. The quantitative results show that most temperature prediction errors are within 2.5 \textdegree C, with a median absolute error of 0.18 \textdegree C. Furthermore, the model inference latency is around 0.01 s on one NVIDIA V100 GPU, achieving \textit{real-time} inference of the airflow and thermal profiles. The result indicates that transformer-based neural operators are capable of learning the high-dimension airflow and thermal profiles of a large-scale DC from limited simulation data, shedding new light on agile proactive DC thermal management in the era of AI computing with high rack power density. 

%% file: body/sec5-future-work.tex
\section{Future Directions}\label{Future Directions}
Based on the initial result from the case study, we hereafter discuss several future directions to be explored.  
\subsection{Foundation Models for PhyAI}
Although PhyAI demonstrates promising results in predictive modeling, it requires extensive domain knowledge to configure the sophisticated physical simulators and build the real-world 3D scene for a physical DC. Large Language Models (LLMs), trained on massive amounts of text data, can deliver human-level intelligence in computer programming, mathematical reasoning, and complex text processing. Operators can leverage LLMs to help them generate complex 3D scenes or first principle-based dynamics models, reducing human efforts and bridging knowledge gaps. Beyond textual knowledge, NVIDIA Cosmos \cite{Cosmos} emerges as a physics-aware foundation model tailored for synthesizing realistic physical environments. By integrating multi-modal data and domain-specific physical priors, Cosmos further ensures generated scenes adhere to conservation laws, physics constraints, and boundary conditions. By embedding physics-aware inductive biases, Cosmos enhances PhyAI’s predictive robustness, scaling simulations to complex real-world scenarios.

\subsection{Physics-informed Training and Adaptation}
Recently, operator learning, an emerging PIML paradigm, has been widely adopted in building fast and accurate surrogate models for PDE-governed tasks as they directly learn input-output mapping and achieve real-time inference of high-dimensional PDE solutions. However, it typically requires hundreds or thousands of solution instances to learn sophisticated input-output mapping for PDE-governed tasks. It incurs huge costs because obtaining solutions via traditional numerical solvers is time-consuming and computationally intensive. In the case study, we conducted an initial study to incorporate the server-level mass and energy conservation into the loss function to constrain the learning process to facilitate convergence. In the future, it would be interesting to explore physics-informed operator learning that injects explicit PDE constraints into the learning objective to improve the sample efficiency or even simulation data-free training. For example, the operator learning model can be trained directly with the PDE residuals similar to the canonical parametric physics-informed neural network without the supervision of the simulation data. In addition, at the inference phase, the known boundary conditions can be leveraged as the objective to adapt the trained model to specific boundary conditions, making the simulation result more physically plausible. 

\subsection{Sample-Efficient Scenario Generation}
Training a PhyAI model requires running large-scale physical simulations multiple times, which takes a long time and consumes significant computing resources. For example, in the case study presented in this article, the parameter dimension is in the order of hundreds, even thousands. To sufficiently sample the entire parameter space, thousands of samples are expected, with each simulation taking several minutes to hours. In this case study, we have proposed a domain decomposition-based approach to alleviate the difficulty in sampling over the high dimensional parameter space. In the future, advanced design of experiment approaches can be explored to reduce the sampling efforts while preserving model performance. 

\subsection{Quantum Computing for Accelerated Physical Simulation}
The integration of quantum computing into physical simulation represents a paradigm shift for solving classically intractable problems in physical simulation. Quantum algorithms like the Variational Quantum Eigensolver (VQE) have demonstrated polynomial-to-exponential speedups in simulating many-body quantum systems, particularly for electronic structure calculations \footnote{https://learning.quantum.ibm.com/tutorial/variational-quantum-eigensolver}. Besides, quantum computing holds transformative potential for accelerating costly CFD simulation, particularly in solving high-dimensional Navier-Stokes equations and turbulence modeling, where classical methods face prohibitive computational costs. Quantum algorithms, such as quantum linear solvers (e.g., Harrow-Hassidim-Lloyd, HHL), promise \textit{exponential speedups} for large-scale sparse linear systems inherent in CFD simulation \cite{syamlal2024quantumCFD}. By leveraging quantum computing technology, it is possible to generate rich synthetic data to train a powerful real-time predictive AI surrogate model under reasonable resource and time budgets, facilitating practical deployment of the PhyAI system to advance future DC operations. 

\subsection{Differentiable Simulation for Prescriptive Analysis}
Since the AI surrogate model is end-to-end differential, exploiting it when designing decision-making agents is also interesting. A possible solution for this is the nonlinear MPC. The AI surrogate model simulates the complex system dynamics while providing the gradient of the performance metric w.r.t. the control inputs or the planned actions. Thus, efficient gradient-based optimization can be employed to derive optimal action trajectories while guaranteeing that DC operation constraints are satisfied. It provides more transparency than the black box DRL agent as the operational constraints can be clearly defined and satisfied under the framework of MPC, mitigating the risk-averse mindset of DC operators. It should be noted that although the gradient can be obtained, the underlying constrained programming is usually difficult to solve due to its nonlinearity and non-convexity. Thus, advanced numerical optimization algorithms need to be adopted to efficiently solve the program in real-time. 

%% file: body/sec6-conclusion.tex
\section{Conclusions}\label{Conclusions}
In this article, we present the concept of Physical AI and its applications in autonomous DC operations. Specifically, we start with the discussion of PhyAI by demonstrating its three building blocks. Subsequently, we present the proposed PhyAI framework for AI-native DC operations. The potential applications of PhyAI are categorized into predictive and prescriptive analysis and discussed accordingly. Finally, we perform a case study, in which we demonstrate how a physics-informed model predicts the temperature and velocity profiles in a large-scale production DC. We show that the model can obtain comparable results with the high-fidelity physical simulator while achieving $10^5$ acceleration and inferring the high-dimensional temperature and velocity fields in real-time.

%% file: main.bbl
\begin{thebibliography}{10}
\providecommand{\url}[1]{#1}
\csname url@samestyle\endcsname
\providecommand{\newblock}{\relax}
\providecommand{\bibinfo}[2]{#2}
\providecommand{\BIBentrySTDinterwordspacing}{\spaceskip=0pt\relax}
\providecommand{\BIBentryALTinterwordstretchfactor}{4}
\providecommand{\BIBentryALTinterwordspacing}{\spaceskip=\fontdimen2\font plus
\BIBentryALTinterwordstretchfactor\fontdimen3\font minus \fontdimen4\font\relax}
\providecommand{\BIBforeignlanguage}[2]{{%
\expandafter\ifx\csname l@#1\endcsname\relax
\typeout{** WARNING: IEEEtran.bst: No hyphenation pattern has been}%
\typeout{** loaded for the language `#1'. Using the pattern for}%
\typeout{** the default language instead.}%
\else
\language=\csname l@#1\endcsname
\fi
#2}}
\providecommand{\BIBdecl}{\relax}
\BIBdecl

\bibitem{hyperscale}
\BIBentryALTinterwordspacing
``Hyperscale data center – what they are and how they work,'' 2023. [Online]. Available: \url{https://www.parallels.com/blogs/ras/hyperscale-data-center/}
\BIBentrySTDinterwordspacing

\bibitem{iea2024}
\BIBentryALTinterwordspacing
``Electricity 2024: Analysis and forecast to 2026,'' 2024. [Online]. Available: \url{https://iea.blob.core.windows.net/assets/18f3ed24-4b26-4c83-a3d2-8a1be51c8cc8/Electricity2024-Analysisandforecastto2026.pdf}
\BIBentrySTDinterwordspacing

\bibitem{gpu-dc}
\BIBentryALTinterwordspacing
``https://www.datacenterdynamics.com/en/opinions/ai-and-data-centers-why-ai-is-so-resource-hungry/.'' [Online]. Available: \url{https://serverlift.com/blog/data-centers-continue-to-see-rack-densities-increase/}
\BIBentrySTDinterwordspacing

\bibitem{price-B200}
\BIBentryALTinterwordspacing
``Nvidia's jensen huang says blackwell gpu to cost $30,000 - $40,000, later clarifies that pricing will vary as they won't sell just the chip,'' 2024. [Online]. Available: \url{https://www.tomshardware.com/pc-components/gpus/nvidias-jensen-huang-says-blackwell-gpu-to-cost-dollar30000-dollar40//000-later-clarifies-that-pricing-will-vary-as-they-wont-sell-just-the-chip?utm_source=chatgpt.com}
\BIBentrySTDinterwordspacing

\bibitem{reliability}
\BIBentryALTinterwordspacing
``Cost of data center outages,'' 2016. [Online]. Available: \url{https://planetaklimata.com.ua/instr/Liebert_Hiross/Cost_of_Data_Center_Outages_2016_Eng.pdf}
\BIBentrySTDinterwordspacing

\bibitem{maturity}
\BIBentryALTinterwordspacing
``The evolving data center management maturity model, a quick update,'' 2019. [Online]. Available: \url{https://journal.uptimeinstitute.com/the-evolving-data-center-management-maturity-model-a-quick-update/}
\BIBentrySTDinterwordspacing

\bibitem{mohapatra2023large}
R.~Mohapatra, A.~Coursey, and S.~Sengupta, ``Large-scale end-of-life prediction of hard disks in distributed datacenters,'' in \emph{2023 ieee international conference on smart computing (smartcomp)}.\hskip 1em plus 0.5em minus 0.4em\relax IEEE, 2023, pp. 261--266.

\bibitem{mao2019learning}
H.~Mao, M.~Schwarzkopf, S.~B. Venkatakrishnan, Z.~Meng, and M.~Alizadeh, ``Learning scheduling algorithms for data processing clusters,'' in \emph{Proceedings of the ACM special interest group on data communication}, 2019, pp. 270--288.

\bibitem{LyaSafe2023}
\BIBentryALTinterwordspacing
Z.~Cao, R.~Wang, X.~Zhou, and Y.~Wen, ``Toward model-assisted safe reinforcement learning for data center cooling control: A lyapunov-based approach,'' in \emph{Proceedings of the 14th ACM International Conference on Future Energy Systems}, ser. e-Energy '23.\hskip 1em plus 0.5em minus 0.4em\relax New York, NY, USA: Association for Computing Machinery, 2023, p. 333–346. [Online]. Available: \url{https://doi.org/10.1145/3575813.3597343}
\BIBentrySTDinterwordspacing

\bibitem{wang2022toward}
R.~Wang, X.~Zhang, X.~Zhou, Y.~Wen, and R.~Tan, ``Toward physics-guided safe deep reinforcement learning for green data center cooling control,'' in \emph{ACM/IEEE 13th Int. Conf. Cyber-Phys. Syst.}\hskip 1em plus 0.5em minus 0.4em\relax IEEE, 2022, pp. 159--169.

\bibitem{uptime-2022}
D.~Jacqueline, B.~Daniel, L.~Andy, R.~Owen, and S.~Max, ``Uptime institute global data center survey 2022,'' 2022.

\bibitem{PhysicalAI}
\BIBentryALTinterwordspacing
``What is physical ai?'' 2025. [Online]. Available: \url{https://www.nvidia.com/en-sg/glossary/generative-physical-ai/}
\BIBentrySTDinterwordspacing

\bibitem{Omniverse}
\BIBentryALTinterwordspacing
``Omniverse.'' [Online]. Available: \url{https://www.nvidia.com/en-sg/omniverse/}
\BIBentrySTDinterwordspacing

\bibitem{PhysicsNeMo}
\BIBentryALTinterwordspacing
``Physicsnemo.'' [Online]. Available: \url{https://developer.nvidia.com/physicsnemo}
\BIBentrySTDinterwordspacing

\bibitem{karniadakis2021physics}
G.~E. Karniadakis, I.~G. Kevrekidis, L.~Lu, P.~Perdikaris, S.~Wang, and L.~Yang, ``Physics-informed machine learning,'' \emph{Nat. Rev. Phys.}, vol.~3, no.~6, pp. 422--440, 2021.

\bibitem{franccois2018introduction}
V.~Fran{\c{c}}ois-Lavet, P.~Henderson, R.~Islam, M.~G. Bellemare, J.~Pineau \emph{et~al.}, ``An introduction to deep reinforcement learning,'' \emph{Foundations and Trends{\textregistered} in Machine Learning}, vol.~11, no. 3-4, pp. 219--354, 2018.

\bibitem{hewing2020learning}
L.~Hewing, K.~P. Wabersich, M.~Menner, and M.~N. Zeilinger, ``Learning-based model predictive control: Toward safe learning in control,'' \emph{Annual Review of Control, Robotics, and Autonomous Systems}, vol.~3, no.~1, pp. 269--296, 2020.

\bibitem{Genesis}
\BIBentryALTinterwordspacing
``Genesis.'' [Online]. Available: \url{https://genesis-world.readthedocs.io/en/latest/index.html}
\BIBentrySTDinterwordspacing

\bibitem{andreoli2025cloudsim}
R.~Andreoli, J.~Zhao, T.~Cucinotta, and R.~Buyya, ``Cloudsim 7g: An integrated toolkit for modeling and simulation of future generation cloud computing environments,'' \emph{Software: Practice and Experience}, 2025.

\bibitem{CadenceDesignCenterDCDigitalTwin}
\BIBentryALTinterwordspacing
``Cadence reality digital twin platform.'' [Online]. Available: \url{https://www.cadence.com/en_US/home/tools/reality-digital-twin.html}
\BIBentrySTDinterwordspacing

\bibitem{jasak2007openfoam}
H.~Jasak, A.~Jemcov, Z.~Tukovic \emph{et~al.}, ``Openfoam: A c++ library for complex physics simulations,'' in \emph{International workshop on coupled methods in numerical dynamics}, vol. 1000.\hskip 1em plus 0.5em minus 0.4em\relax Dubrovnik, Croatia), 2007, pp. 1--20.

\bibitem{crawley2001energyplus}
D.~B. Crawley, L.~K. Lawrie, F.~C. Winkelmann, W.~F. Buhl, Y.~J. Huang, C.~O. Pedersen, R.~K. Strand, R.~J. Liesen, D.~E. Fisher, M.~J. Witte \emph{et~al.}, ``Energyplus: creating a new-generation building energy simulation program,'' \emph{Energy and buildings}, vol.~33, no.~4, pp. 319--331, 2001.

\bibitem{cao2022reducio}
Z.~Cao, R.~Wang, X.~Zhou, and Y.~Wen, ``Reducio: Model reduction for data center predictive digital twins via physics-guided machine learning,'' in \emph{Proc. 7th ACM Int. Conf. Syst. Energy-Efficient Buildings Cities Transp.}\hskip 1em plus 0.5em minus 0.4em\relax ACM, 2022, pp. 1--10.

\bibitem{wang2020kalibre}
R.~Wang, X.~Zhou, L.~Dong, Y.~Wen, R.~Tan, L.~Chen, G.~Wang, and F.~Zeng, ``Kalibre: Knowledge-based neural surrogate model calibration for data center digital twins,'' in \emph{Proc. 7th ACM Int. Conf. Syst. Energy-Efficient Buildings Cities Transp.}, 2020, pp. 200--209.

\bibitem{Cao2023Adaptive}
Z.~Cao, R.~Wang, X.~Zhou, R.~Tan, Y.~Wen, Y.~Yan, and Z.~Wang, ``Adaptive capacity provisioning for carbon-aware data centers: a digital twin-based approach,'' \emph{IEEE Transactions on Sustainable Computing}, pp. 1--15, 2025.

\bibitem{Zhou2023JointIT}
X.~Zhou, R.~Wang, Y.~Wen, and R.~Tan, ``Joint it-facility optimization for green data centers via deep reinforcement learning,'' \emph{IEEE Network}, vol.~35, no.~6, pp. 255--262, 2021.

\bibitem{lazic2018data}
N.~Lazic, C.~Boutilier, T.~Lu, E.~Wong, B.~Roy, M.~Ryu, and G.~Imwalle, ``Data center cooling using model-predictive control,'' in \emph{Adv. Neural Inf. Process. Syst.}, 2018, pp. 3814--3823.

\bibitem{hao2023gnot}
Z.~Hao, Z.~Wang, H.~Su, C.~Ying, Y.~Dong, S.~Liu, Z.~Cheng, J.~Song, and J.~Zhu, ``Gnot: A general neural operator transformer for operator learning,'' in \emph{International Conference on Machine Learning}.\hskip 1em plus 0.5em minus 0.4em\relax PMLR, 2023, pp. 12\,556--12\,569.

\bibitem{tancik2020fourier}
M.~Tancik, P.~Srinivasan, B.~Mildenhall, S.~Fridovich-Keil, N.~Raghavan, U.~Singhal, R.~Ramamoorthi, J.~Barron, and R.~Ng, ``Fourier features let networks learn high frequency functions in low dimensional domains,'' \emph{Advances in neural information processing systems}, vol.~33, pp. 7537--7547, 2020.

\bibitem{Cosmos}
\BIBentryALTinterwordspacing
``Cosmos.'' [Online]. Available: \url{https://www.nvidia.com/en-sg/ai/cosmos/}
\BIBentrySTDinterwordspacing

\bibitem{syamlal2024quantumCFD}
M.~Syamlal, C.~Copen, M.~Takahashi, and B.~Hall, ``Computational fluid dynamics on quantum computers,'' in \emph{AIAA AVIATION FORUM AND ASCEND 2024}, 2024, p. 3534.

\end{thebibliography}
